\documentclass{article}
\usepackage[preprint]{spconf}
\usepackage{amsmath,graphicx}
\usepackage{booktabs}
\usepackage{colortbl}
\usepackage{mathtools}
\usepackage{cleveref}
\usepackage{tikz}
\usepackage[T1]{fontenc}

\crefname{table}{Tab.}{Tabs.}
\crefname{section}{Sec.}{Secs.}
\crefname{figure}{Fig.}{Figs.}
\crefname{equation}{Eq.}{Eqs.}

\newcommand{\skipping}[1]{}
\copyrightnotice{}
               \toappear{\parbox{\textwidth}{ \footnotesize\linespread{1}\selectfont\copyright\ IEEE 2022. Personal use of this material is permitted. Permission from IEEE must be obtained for all other uses, in any current or future media, including reprinting/republishing this material for advertising or promotional purposes, creating new collective works, for resale or redistribution to servers or lists, or reuse of any copyrighted component of this work in other works.}}
\title{Multi-Scale RAFT: Combining Hierarchical Concepts for Learning-based Optical Flow Estimation
}
\name{Azin Jahedi$^1$ \qquad Lukas Mehl$^1$  \qquad Marc Rivinius$^2$  \qquad Andrés Bruhn$^1$\thanks{
The authors thank the Deutsche Forschungsgemeinschaft (DFG, German Research Foundation) -- Project-ID 251654672 -- TRR 161 (B04) for funding and the state of Baden-Württemberg for their support through bwHPC. Moreover, they thank the International Max Planck Research School for Intelligent Systems (IMPRS-IS) for supporting Azin Jahedi. 
}}
\address{
$^1$ Institute for Visualization and Interactive Systems, University of Stuttgart, Germany\\
$^2$ Institute of Information Security, University of Stuttgart, Germany}

\begin{document}
%
\maketitle
\begin{abstract}

Many classical and learning-based optical flow methods rely on hierarchical concepts to improve both accuracy and robustness. However, one of the currently most successful approaches -- RAFT -- hardly exploits such concepts. In this work, we show that multi-scale ideas are still valuable.
More precisely, using RAFT as a baseline, we propose a novel multi-scale neural network that combines several hierarchical concepts within a single estimation framework. These concepts include 
(i) a partially shared coarse-to-fine architecture, (ii) multi-scale features, (iii) a hierarchical cost volume and (iv) a multi-scale multi-iteration loss.
Experiments on MPI Sintel and KITTI clearly demonstrate the
benefits of our approach.
They show not only sub\-stantial improvements compared to RAFT, 
but also  state-of-the-art results -- in particular in non-occluded regions. Code will be available at \small{ \texttt{https://github.com/cv-stuttgart/MS\_RAFT}}.
\end{abstract}
\begin{keywords}
Optical flow, Neural networks, Coarse-to-fine estimation, Multi-scale features, Multi-scale loss 
\end{keywords}
\section{Introduction}
\label{sec:intro}

The estimation of the optical flow is one of the key problems in computer vision. 
To solve this task many approaches rely on matching characteristic image features by minimizing some local or global cost. In this context, hierarchical matching schemes have a long tradition, since they address both accuracy and robustness
\cite{Moravec1979_VisMap,
Anandan1989,
Black1991,
Brox2004,
Xu2011_MDPFlow,Hu2016_CPM}. 
By considering flow initializations from coarser scales,
they not only enable the estimation of larger motion 
but also render the underlying minimization more robust w.r.t.\ local minima.

Meanwhile, hierarchical schemes also became popular in the context of neural networks. For instance, SpyNet 
\cite{Ranjan2017_SPyNet} proposes to use a spatial pyramid network, where the downsampled input frames are warped
by the current estimate and 
iterative updates proceed in a coarse-to-fine manner \cite{Brox2004}.
PWC-Net \cite{Sun2018_PWC} improved this idea by warping image features
instead of input frames and by a joint training of the individual scales using
 a multi-scale loss. Finally, IRR \cite{Hur2019} suggests to share the estimation network across all scales, leading to a more compact architecture which can be trained more efficiently. 

In contrast, RAFT \cite{Teed2020_RAFT} follows a completely different ap- proach.
It performs recurrent updates on a single scale with spatial size $\frac{1}{8}\times (h,w)$
finally interpolating the results.
When performing the updates, however, it relies on a sampling-based hierarchical cost volume, i.e.\ a correlation pyramid with multiple lookup levels.
While the correlation pyramid adds non-local cost information to the matching process, the sampling allows to overcome some
problems of 
warping
\cite{Hofinger2020_IOFPL}.

Given the orthogonal hierarchical concepts considered in PWC-Net / IRR and RAFT, it seems desirable to combine them within a single architecture. Moreover, in this context, it seems also desirable to integrate other hierarchical concepts, e.g.\ regarding the features and regarding the loss.

\medskip

\noindent {\bf Contributions.} Hence, in this work, we propose a multi-scale neural network that builds upon RAFT and combines several established hierarchical concepts within a single approach. In this context, our contributions are fourfold:
(i) We embed the 
recurrent 
estimation in a coarse-to-fine architecture 
-- using a shared sub-network -- 
which enables us to build upon results from coarser scales.
(ii) Besides, similar to feature pyramid networks (FPNs) \cite{Lin2017_FPN}, we propose to use U-Net-style features that allow to exploit multi-scale information also during matching.
(iii) Moreover, when combining our coarse-to-fine architecture with our multi-scale features, we suggest to still use RAFT's correlation pyramid as third hierarchical concept on top, since it provides additional information in terms of non-local costs.
(iv) Finally, we propose a multi-scale multi-iteration loss that relies on a 
novel sample-wise robustification and hence allows for a more robust training. Experiments on Sintel \cite{Butler2012_Sintel} and KITTI \cite{Menze2015_KITTI} demonstrate that our multi-scale efforts pay off. They show clear improvements compared to RAFT and state-of-the-art results on both benchmarks - with a particularly high accuracy in non-occluded regions.

\medskip

\noindent {\bf Related Work.} To the best of our knowledge, there is only the report of Wan {\em et al.\ }\cite{Wan2020_PRAFlow} that seeks to extend RAFT to more than one scale. 
However, in their approach only two scales are investigated, features 
are limited to a bottom-up 
approach (cf.\ \cite{Lin2017_FPN})
and the actual loss is not specified.
Regarding the use of multi-scale features,
the
methods of Saxena {\em et al.\ }\cite{Saxena2019_pwoc}, Rishav {\em et al.\ }\cite{Rishav2021_ResFPN} and Long and Lang \cite {Long2022}
have to be mentioned. 
While 
\cite{Saxena2019_pwoc,Rishav2021_ResFPN} are also based on 
FPN ideas to 
integrate semantically important information from other scales,
they rely on PWC-Net  \cite{Sun2018_PWC} 
and hence neither exploit recurrent updates nor a non-local lookup in the cost volume. 
On the other hand, while 
\cite {Long2022} is 
based on RAFT, it estimates the flow on a single scale thereby using a feature extractor
that 
aims at preserving fine-scale details
via residual downsampling.

\section{Approach}
\label{sec:approach}
Let us now discuss the details of our multi-scale method. In the following, we explain the coarse-to-fine architecture, the U-Net-style feature extractor, the integration of RAFT's correlation pyramid and our multi-scale multi-iteration loss. 

\begin{figure}
    \centering
    \includegraphics[width=0.86\columnwidth]{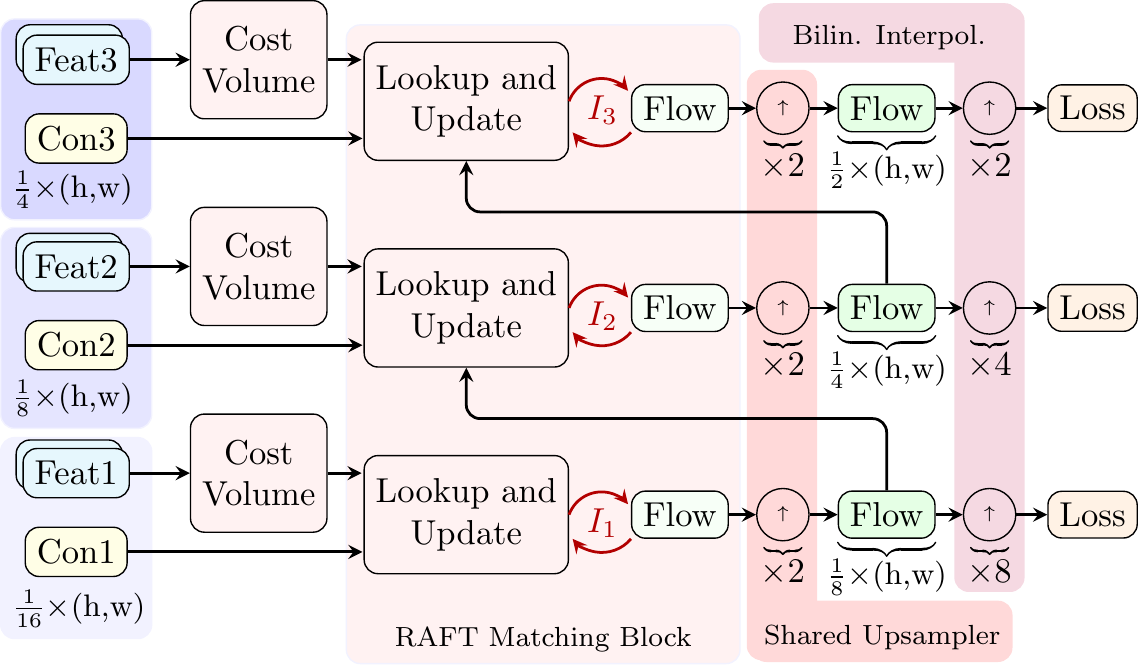}
    \caption{Our coarse-to-fine architecture for 3 scales.} 
    
\label{fig:multi_scale_flow}
\end{figure}

\smallskip

\noindent {\bf Coarse-to-fine Architecture.}
 \cref{fig:multi_scale_flow} makes the coarse-to-fine\linebreak architecture of our
method
explicit.
Inputs at different scales are image features 
from both frames (cyan) and context features 
from the first frame (yellow); see next section.
Starting from the coarsest scale, a cost volume is built from the image features and from it a correlation pyramid is constructed.
Based on the current estimate 
-- which is zero at the beginning --
candidate correlations are selected from the correlation pyramid and processed together with the context features and the current flow estimate 
via a recurrent unit that computes the update; see RAFT \cite{Teed2020_RAFT} for details. 
After performing several updates, the flow is upsampled via a $\times 2$ learned convex 
mask
-- similar to the $\times 8$ convex upsampling in \cite{Teed2020_RAFT} --
to the next finer scale, where it serves as initialization. 
This procedure is repeated until the flow is updated at the finest scale $\frac{1}{4}\times(h,w)$, i.e.\ twice the resolution of RAFT.
Finally, the result is interpolated to the original resolution ($\times 2$ convex upsampling followed by $\times 2$ bilinear interpolation).
Note that the 
matching block
and the upsampling mask are shared among scales.

\begin{figure}
    \centering
    \includegraphics[width=0.72\columnwidth]{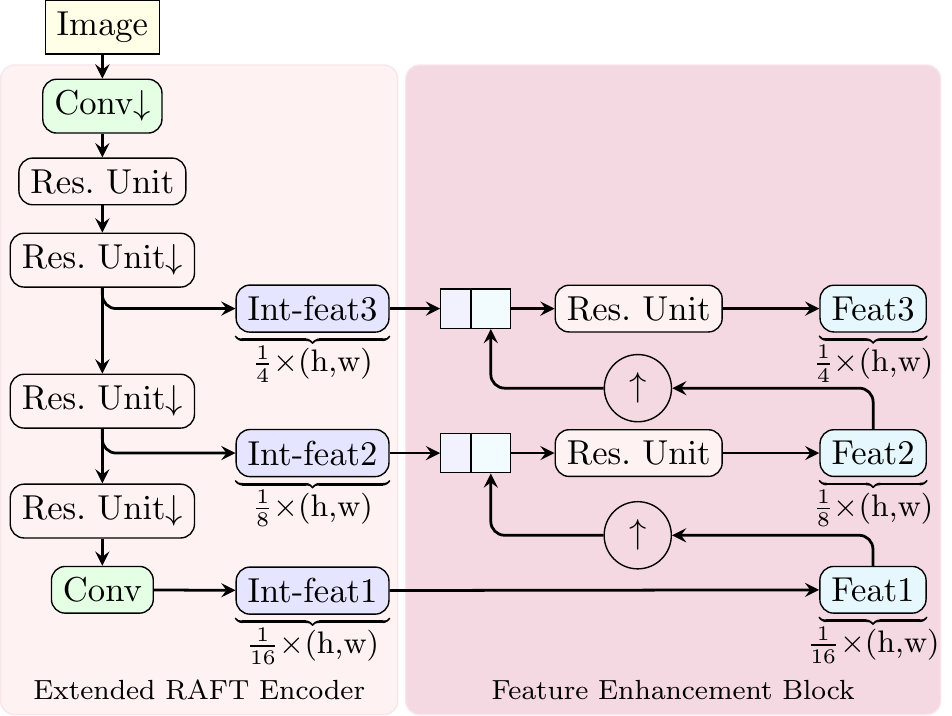}
    \caption{Our U-net-style feature extractor for 3 scales.}
    \label{fig:unet}
\end{figure}

\smallskip

\noindent {\bf U-Net-style Features.}
The architecture of our feature extractor 
is illustrated in \cref{fig:unet}. 
Its left hand-side shows the encoder of RAFT that is extended such that residual units -- one in our example with 3 scales -- have been added to introduce 
features at coarser scales.
Also, features at scale $\frac{1}{4}\times(h,w)$ are con- sidered compared to RAFT.
Unlike 
\cite{Sun2018_PWC, Hur2019, Wan2020_PRAFlow} 
we do not use the direct outputs of such an encoder, except for the coarsest scale.
Instead, in the feature enhancement block -- similar to \cite{Saxena2019_pwoc} -- we semantically enrich our features by gathering information from the previous coarser scale.
To this end, we follow a U-Net-like strategy \cite{Ronneberger2015_UNet} and concatenate the upsampled features of the previous coarser scale (Feat $\!{k\!-\!1}$) and the better localized intermediate features of the current scale (Int-feat $\!{k}$). 
The resulting volume is further processed via a residual unit to aggregate relevant information. 
Note that the same architecture is used to compute the context features. 
However, 
while in case of image features, the number of output channels may vary per scale, in case of context features it must be identical -- to allow the matching block in  \cref{fig:multi_scale_flow} to be shared.

\tikzset{rectstyle/.style={draw=black,densely dotted}}
\tikzset{rectstyle2/.style={draw=white, densely dotted}}
\tikzset{labelstyle/.style={anchor=south west, fill=white, inner sep=1, text opacity=1, fill opacity=0.01, scale=0.35}}

\begin{figure*}[t!]
    \centering
    \begin{tabular}{ccccccccccc}
    \footnotesize{Ground truth / 1st frame} 
    &\!\!\!\!\!\!\!\!\!\!\!\!\!& 
    \footnotesize{Our method} 
    &\!\!\!\!\!\!\!\!\!\!\!\!\!&
    \footnotesize{GMA} 
    &\!\!\!\!\!\!\!\!\!\!\!\!\!& 
    \footnotesize{RFPM} 
    &\!\!\!\!\!\!\!\!\!\!\!\!\!& 
    \footnotesize{SeparableFlow} 
    &\!\!\!\!\!\!\!\!\!\!\!\!\!&
    \footnotesize{RAFT}
    \\    
    \begin{tikzpicture}
    \coordinate (r11A) at (1.04,-0.02);
    \coordinate (r11B) at (1.95,-0.92);
    \coordinate (r12A) at (2.1,-0.15);
    \coordinate (r12B) at (2.6,-0.65);
    \coordinate (r21A) at (0.7,-0.72);
    \coordinate (r21B) at (1.1,-0.7);
    \coordinate (r22A) at (1.8,-0.72);
    \coordinate (r22B) at (2.2,-0.35);
    \coordinate (r31A) at (0.05,-0.03);
    \coordinate (r31B) at (0.8,-0.5);
    \coordinate (r32A) at (1.9,-0.1);
    \coordinate (r32B) at (2.75,-0.55);
    \draw (0, 0) node[inner sep=0,anchor=north west] (img) {
    \includegraphics[width=0.155\textwidth]{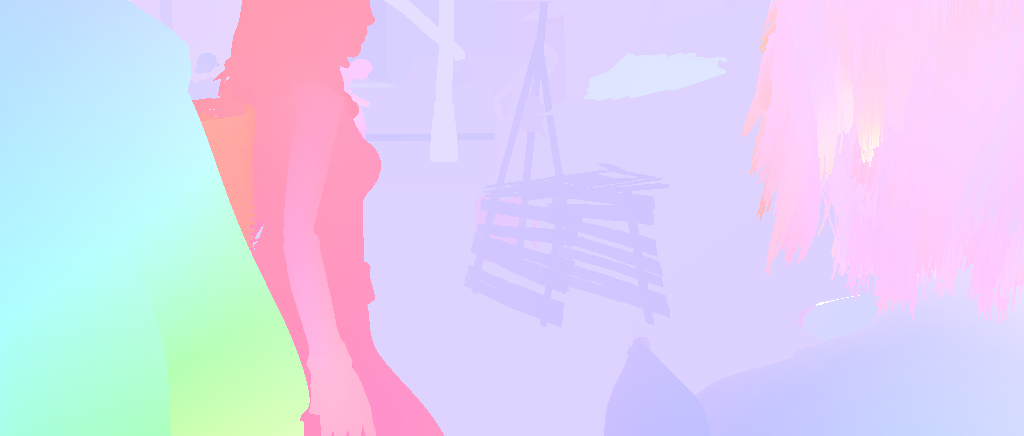}
    }; \draw[rectstyle] (r11A) rectangle (r11B); \draw[rectstyle] (r12A) rectangle (r12B);
    \draw (img.south west) node[labelstyle, yshift=3, xshift=2] {\phantom{EPE 0.000}};\end{tikzpicture}
    &\!\!\!\!\!\!\!\!\!\!\!\!\!&
    \begin{tikzpicture} \draw (0, 0) node[inner sep=0,anchor=north west] (img) {
	\includegraphics[width=0.155\textwidth]{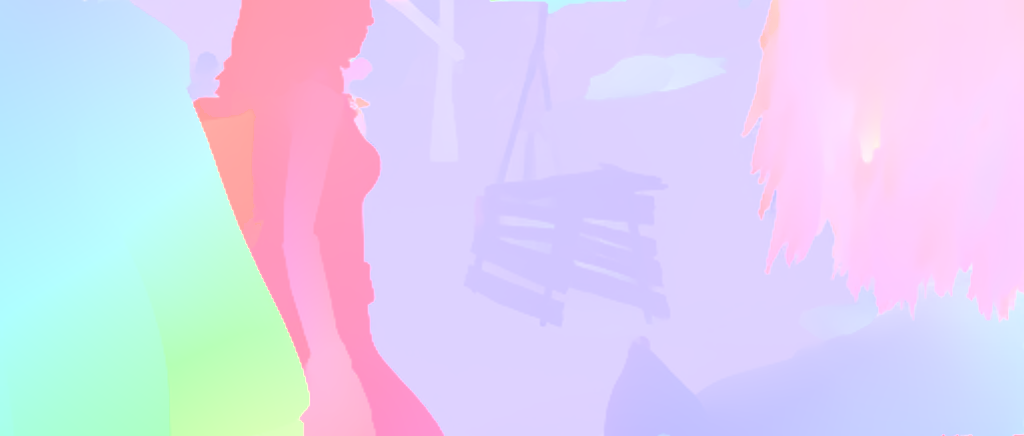}
	}; \draw[rectstyle] (r11A) rectangle (r11B); \draw[rectstyle] (r12A) rectangle (r12B);
	\draw (img.south west) node[labelstyle, yshift=3, xshift=2] {EPE 0.452};
	\end{tikzpicture}
	&\!\!\!\!\!\!\!\!\!\!\!\!\!&
	\begin{tikzpicture} \draw (0, 0) node[inner sep=0,anchor=north west] (img) {
	\includegraphics[width=0.155\textwidth]{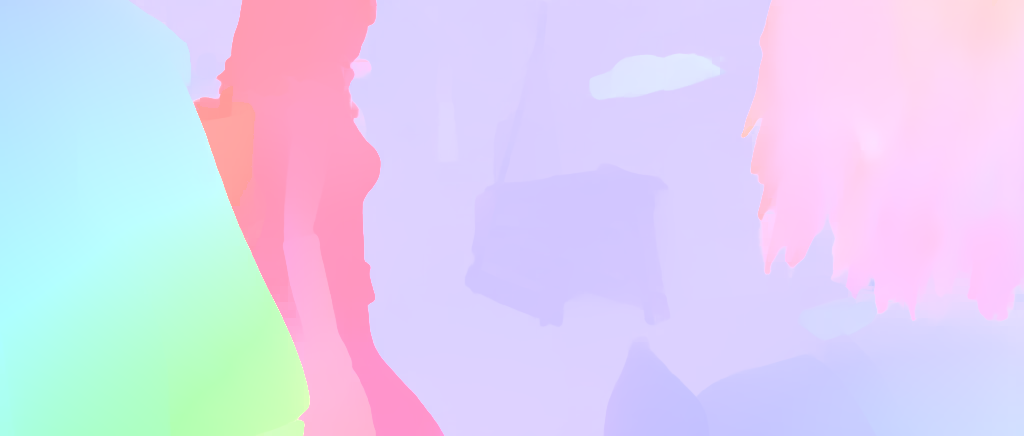}
	}; \draw[rectstyle] (r11A) rectangle (r11B); \draw[rectstyle] (r12A) rectangle (r12B);
	\draw (img.south west) node[labelstyle, yshift=3, xshift=2] {EPE 0.624};
	\end{tikzpicture}
	&\!\!\!\!\!\!\!\!\!\!\!\!\!&
	\begin{tikzpicture} \draw (0, 0) node[inner sep=0,anchor=north west] (img) {
	\includegraphics[width=0.155\textwidth]{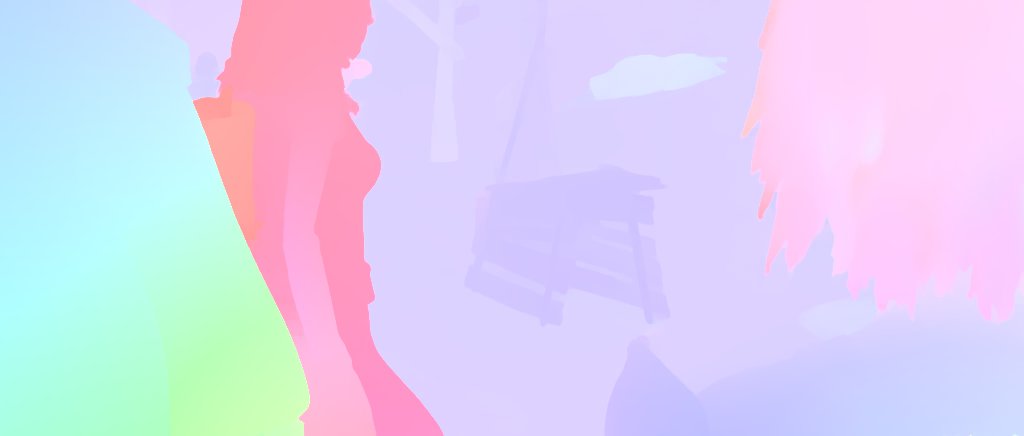}
	}; \draw[rectstyle] (r11A) rectangle (r11B); \draw[rectstyle] (r12A) rectangle (r12B);
	\draw (img.south west) node[labelstyle, yshift=3, xshift=2] {EPE 0.528};
	\end{tikzpicture}
	&\!\!\!\!\!\!\!\!\!\!\!\!\!&
	\begin{tikzpicture} \draw (0, 0) node[inner sep=0,anchor=north west] (img) {
	\includegraphics[width=0.155\textwidth]{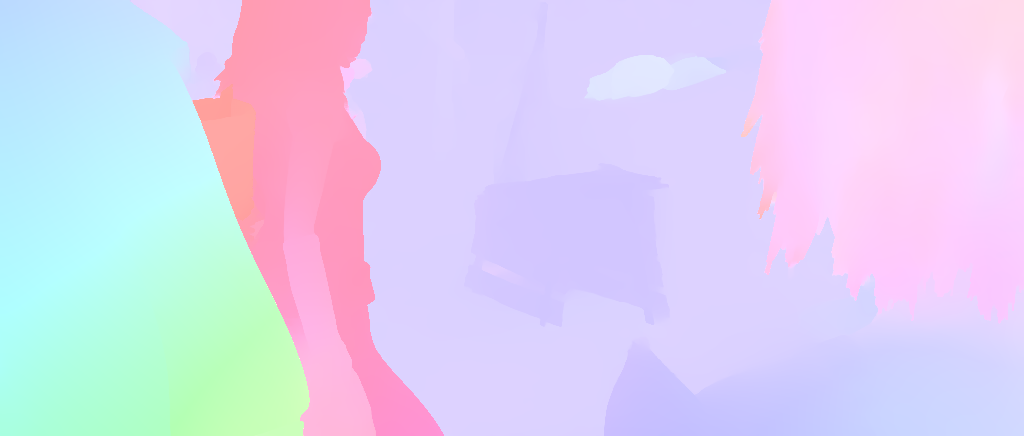}
	}; \draw[rectstyle] (r11A) rectangle (r11B); \draw[rectstyle] (r12A) rectangle (r12B);
	\draw (img.south west) node[labelstyle, yshift=3, xshift=2] {EPE 0.664};
	\end{tikzpicture}
	&\!\!\!\!\!\!\!\!\!\!\!\!\!&
	\begin{tikzpicture} \draw (0, 0) node[inner sep=0,anchor=north west] (img) {
	\includegraphics[width=0.155\textwidth]{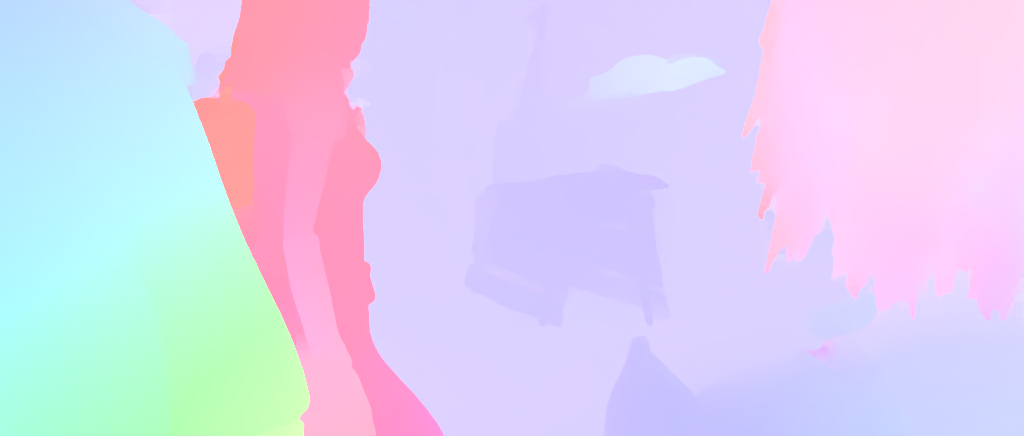}
	}; \draw[rectstyle] (r11A) rectangle (r11B); \draw[rectstyle] (r12A) rectangle (r12B);
	\draw (img.south west) node[labelstyle, yshift=3, xshift=2] {EPE 0.679};
	\end{tikzpicture}
	\\[-0.07cm]
	\begin{tikzpicture} \draw (0, 0) node[inner sep=0,anchor=north west] (img) {
	\includegraphics[width=0.155\textwidth]{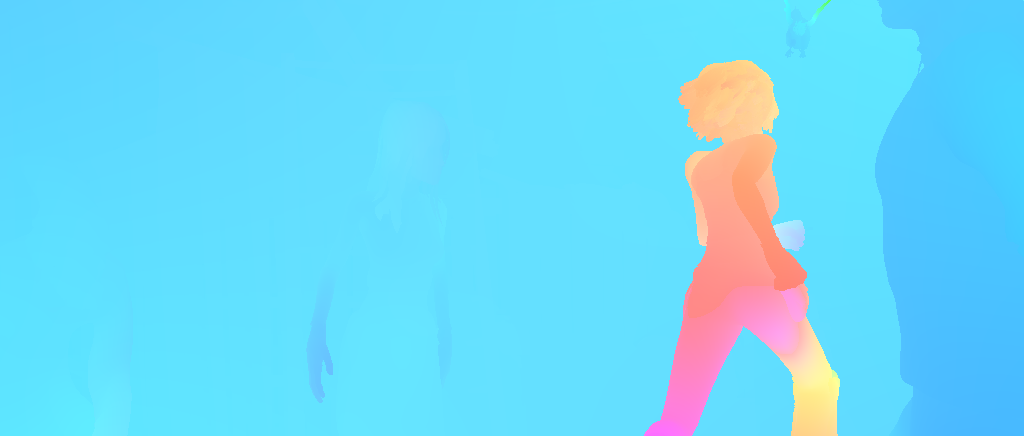}
	}; \draw[rectstyle] (r22A) rectangle (r22B);
	\draw (img.south west) node[labelstyle, yshift=3, xshift=2] {\phantom{EPE 0.000}};\end{tikzpicture}
	&\!\!\!\!\!\!\!\!\!\!\!\!\!&
	\begin{tikzpicture} \draw (0, 0) node[inner sep=0,anchor=north west] (img) {
	\includegraphics[width=0.155\textwidth]{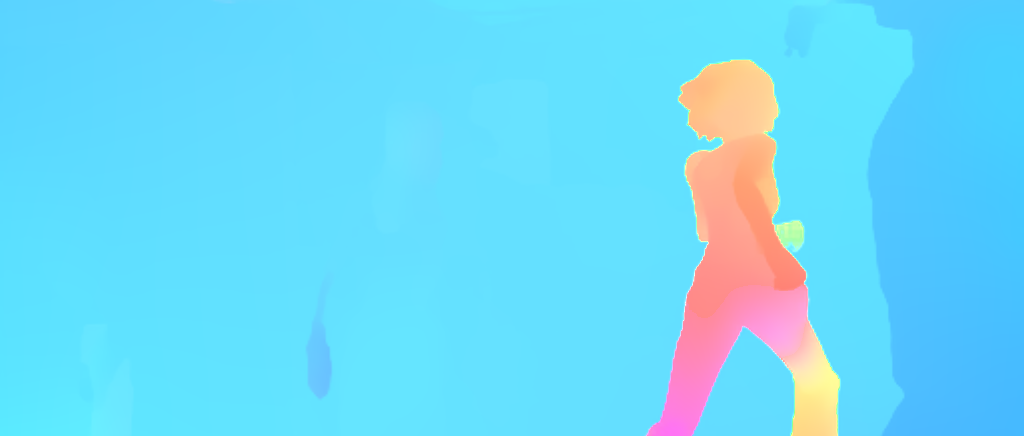}
	}; \draw[rectstyle] (r22A) rectangle (r22B);
	\draw (img.south west) node[labelstyle, yshift=3, xshift=2] {EPE 1.133};
	\end{tikzpicture}
	&\!\!\!\!\!\!\!\!\!\!\!\!\!&
	\begin{tikzpicture} \draw (0, 0) node[inner sep=0,anchor=north west] (img) {
	\includegraphics[width=0.155\textwidth]{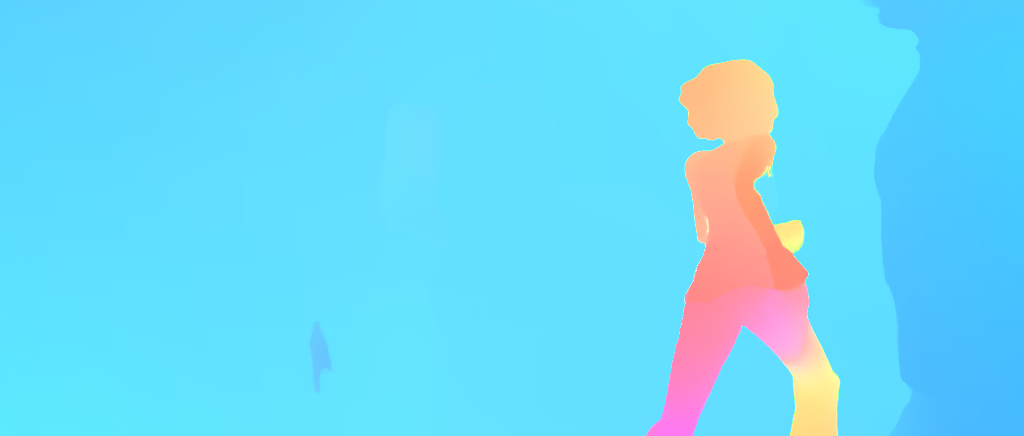}
	}; \draw[rectstyle] (r22A) rectangle (r22B);
	\draw (img.south west) node[labelstyle, yshift=3, xshift=2] {EPE 1.212};
	\end{tikzpicture}
	&\!\!\!\!\!\!\!\!\!\!\!\!\!&
	\begin{tikzpicture} \draw (0, 0) node[inner sep=0,anchor=north west] (img) {
	\includegraphics[width=0.155\textwidth]{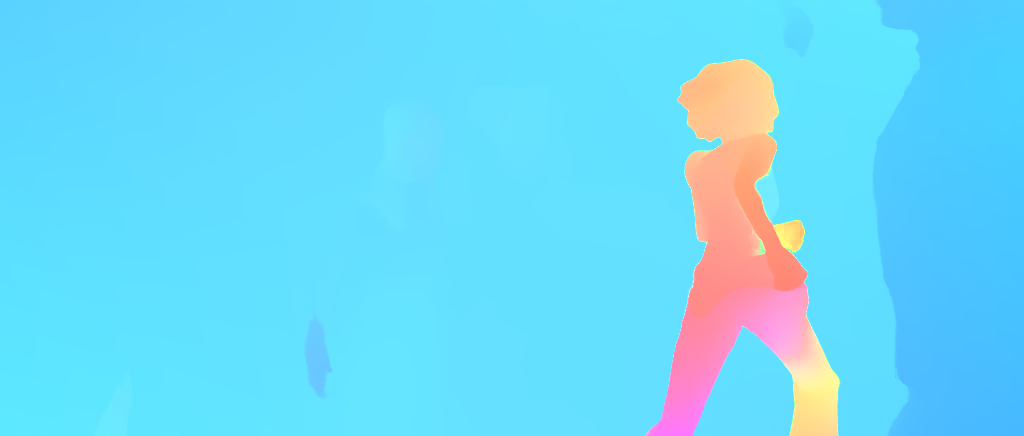}
	}; \draw[rectstyle] (r22A) rectangle (r22B);
	\draw (img.south west) node[labelstyle, yshift=3, xshift=2] {EPE 1.382};
	\end{tikzpicture}
	&\!\!\!\!\!\!\!\!\!\!\!\!\!&
	\begin{tikzpicture} \draw (0, 0) node[inner sep=0,anchor=north west] (img) {
	\includegraphics[width=0.155\textwidth]{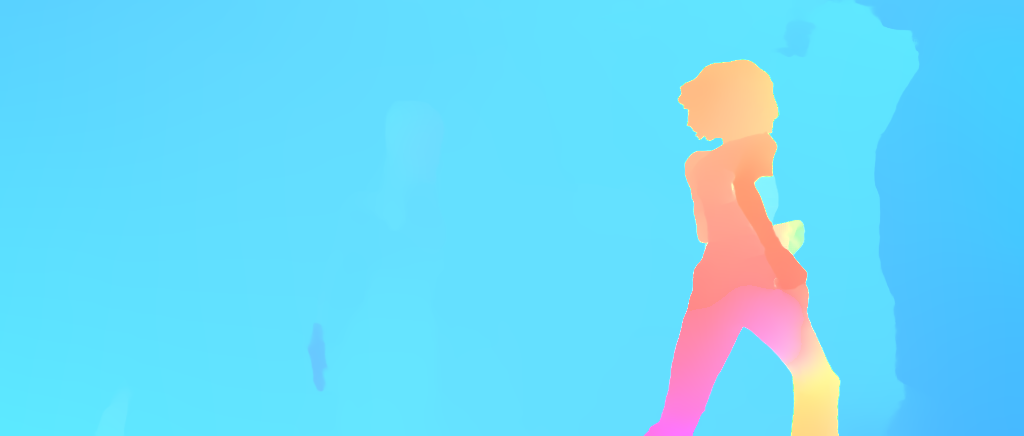}
	}; \draw[rectstyle] (r22A) rectangle (r22B);
	\draw (img.south west) node[labelstyle, yshift=3, xshift=2] {EPE 1.261};
	\end{tikzpicture}
	&\!\!\!\!\!\!\!\!\!\!\!\!\!&
	\begin{tikzpicture} \draw (0, 0) node[inner sep=0,anchor=north west] (img) {
	\includegraphics[width=0.155\textwidth]{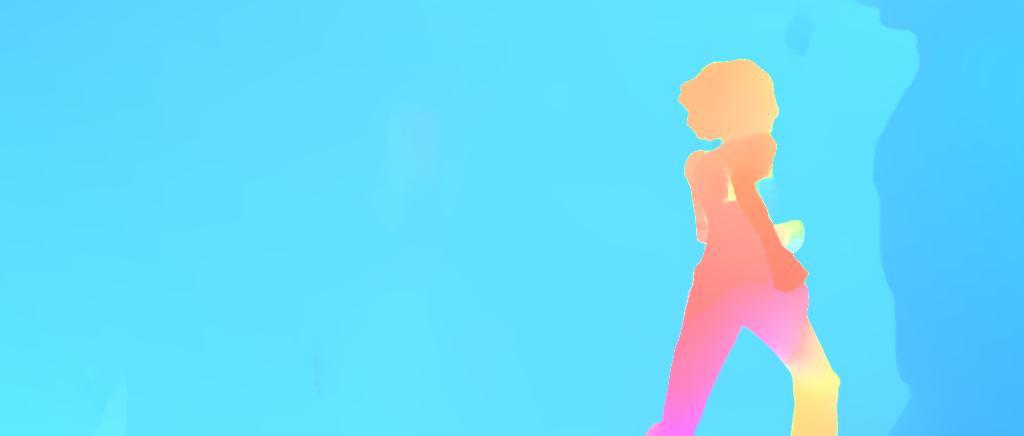}
	}; \draw[rectstyle] (r22A) rectangle (r22B);
	\draw (img.south west) node[labelstyle, yshift=3, xshift=2] {EPE 1.445};
	\end{tikzpicture}
	\\[-0.07cm] 
	\begin{tikzpicture} \draw (0, 0) node[inner sep=0,anchor=north west] (img) {
	\includegraphics[width=0.155\textwidth]{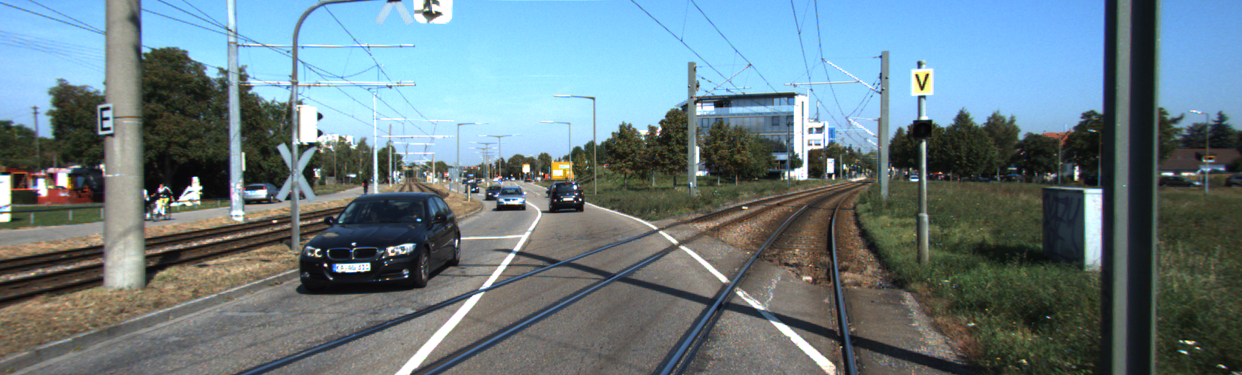}
	}; \draw[rectstyle2] (r31A) rectangle (r31B); \draw[rectstyle2] (r32A) rectangle (r32B);
	\draw (img.south west) node[labelstyle, yshift=3, xshift=2] {\phantom{Fl 0.000}};\end{tikzpicture}
	&\!\!\!\!\!\!\!\!\!\!\!\!\!&
	\begin{tikzpicture} \draw (0, 0) node[inner sep=0,anchor=north west] (img) {
	\includegraphics[width=0.155\textwidth]{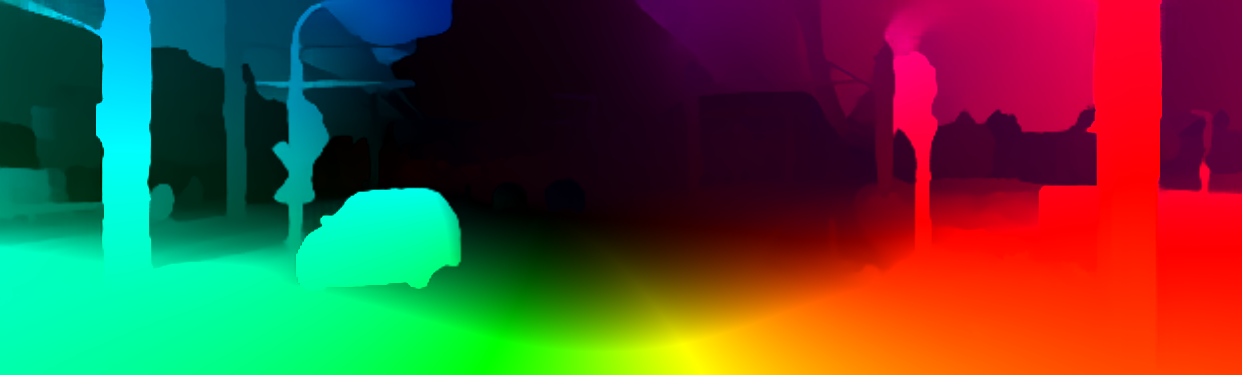}
	}; \draw[rectstyle2] (r31A) rectangle (r31B); \draw[rectstyle2] (r32A) rectangle (r32B);
	\draw (img.south west) node[labelstyle, yshift=3, xshift=2] {Fl 11.34};
	\end{tikzpicture}
	&\!\!\!\!\!\!\!\!\!\!\!\!\!&
	\begin{tikzpicture} \draw (0, 0) node[inner sep=0,anchor=north west] (img) {
	\includegraphics[width=0.155\textwidth]{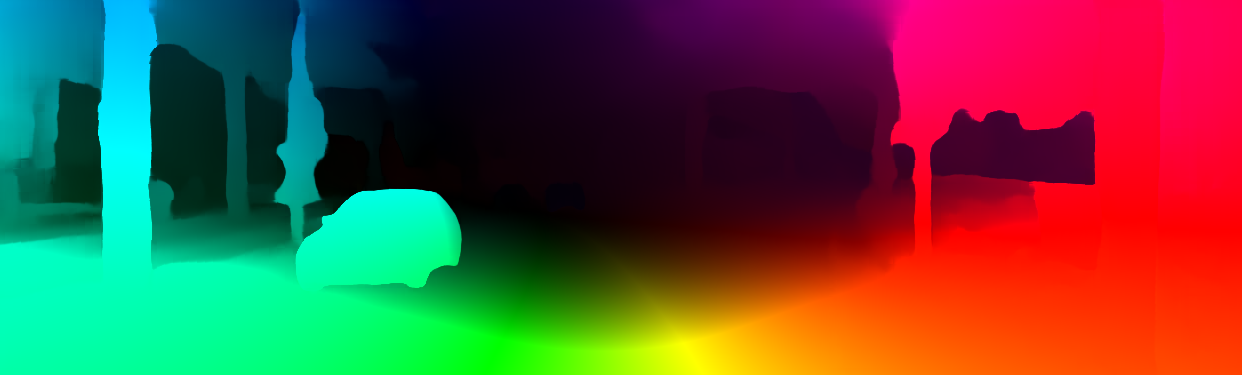}
	}; \draw[rectstyle2] (r31A) rectangle (r31B); \draw[rectstyle2] (r32A) rectangle (r32B);
	\draw (img.south west) node[labelstyle, yshift=3, xshift=2] {Fl 11.63};
	\end{tikzpicture}
	&\!\!\!\!\!\!\!\!\!\!\!\!\!&
	\begin{tikzpicture} \draw (0, 0) node[inner sep=0,anchor=north west] (img) {
    \includegraphics[width=0.155\textwidth]{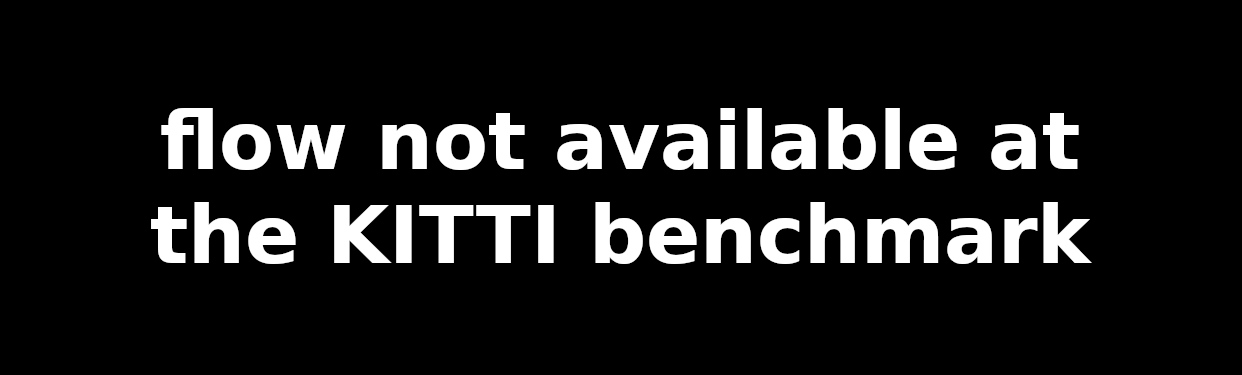}
    }; 
	\draw (img.south west) node[labelstyle, yshift=3, xshift=2] {\phantom{Fl 0.00}};
	\end{tikzpicture}
    &\!\!\!\!\!\!\!\!\!\!\!\!\!&
    \begin{tikzpicture} \draw (0, 0) node[inner sep=0,anchor=north west] (img) {
	\includegraphics[width=0.155\textwidth]{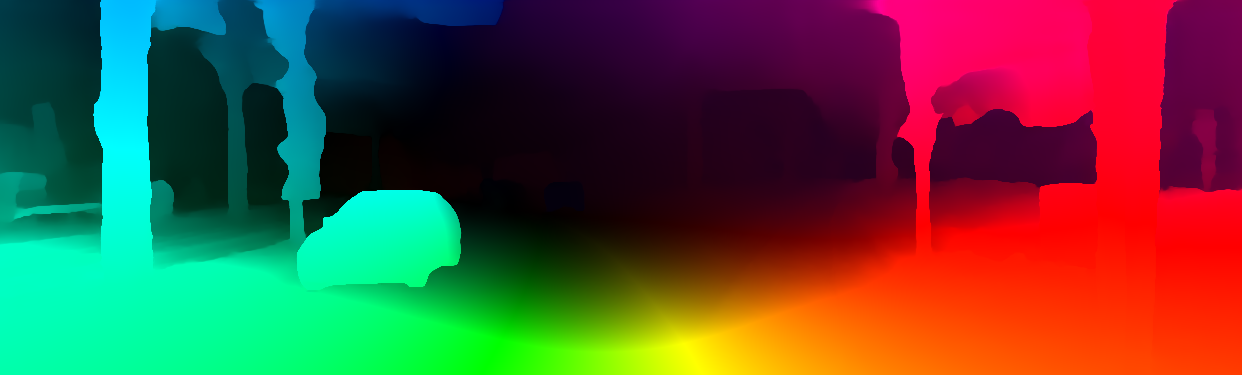}
	}; \draw[rectstyle2] (r31A) rectangle (r31B); \draw[rectstyle2] (r32A) rectangle (r32B);
	\draw (img.south west) node[labelstyle, yshift=3, xshift=2] {Fl 12.49};
	\end{tikzpicture}
	&\!\!\!\!\!\!\!\!\!\!\!\!\!&
	\begin{tikzpicture} \draw (0, 0) node[inner sep=0,anchor=north west] (img) {
	\includegraphics[width=0.155\textwidth]{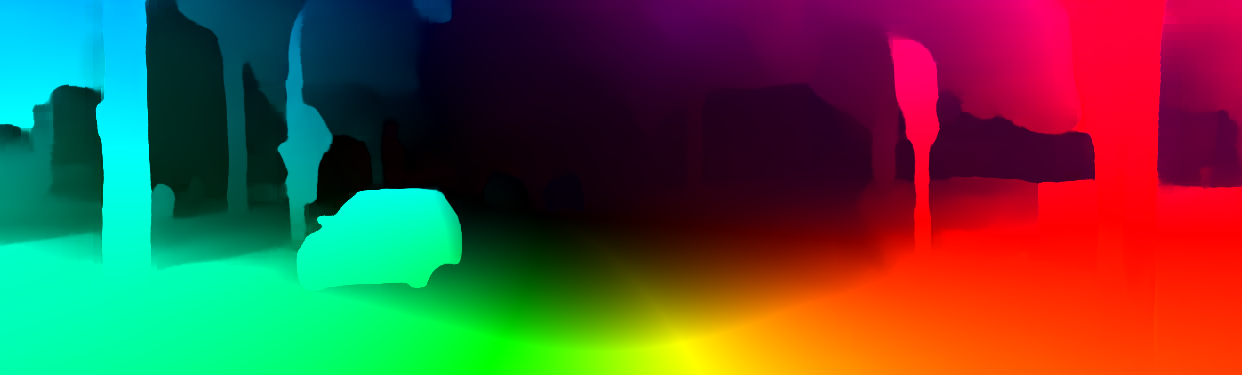}
	}; \draw[rectstyle2] (r31A) rectangle (r31B); \draw[rectstyle2] (r32A) rectangle (r32B);
	\draw (img.south west) node[labelstyle, yshift=3, xshift=2] {Fl 12.57};
	\end{tikzpicture}
    \end{tabular}
    \caption{Computed optical flow for the test set of Sintel Clean and Final (top, middle) and KITTI (bottom), best viewed as PDF.}
    \label{fig:benchmark_results_images} 
\end{figure*}
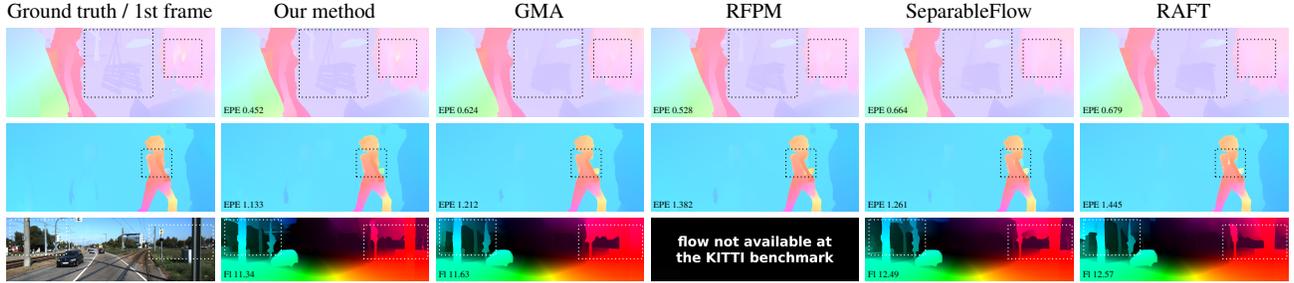

\smallskip 

\noindent {\bf Integration of the Correlation Pyramid.}
While our coarse-to-fine architecture provides flow initializations from coarser scales,
originally it does not consider non-local cost information during the actual matching at the current scale.
Such information, however, may help to escape local minima when performing 
the recurrent updates. 
Thus, RAFT's correlation\linebreak pyramid
complements our coarse-to-fine strategy. 
When combining both concepts, we adjust the number of lookup levels of the correlation pyramid such that, at coarser scales, not too many looked-up candidates lie outside the image domain. This is required, since our method starts from a coarser scale than RAFT and moreover uses a shared update block.

\smallskip

\noindent {\bf Multi-scale Multi-iteration Loss.}
Since in each iteration of each scale, the current flow estimate plays a pivotal role in estimating the flow update, we consider a multi-scale multi-iteration loss combining ideas from \cite{Sun2018_PWC} and \cite{Teed2020_RAFT}.
Let $N_{\mathrm{scales}}$ be the number of resolution scales and $N_{\mathrm{iters}}(s)$ the number of iterations at scale $s$. Then our overall loss is given by
\begin{equation}
\label{msmiloss:sum}
     \mathcal{L} = \sum_{s=1}^{N_{\mathrm{scales}}} \sum_{i=1}^{N_{\mathrm{iters}}(s)} \gamma_{s,i} \cdot L_{s,i}
\end{equation}
where the loss at resolution scale $s$ 
and iteration $i$ 
reads 
\begin{equation}
     \label{msmiloss:pre}
     L_{s,i} = \frac{1}{N_{\mathrm{spb}}}\sum_{\mathrm{m}=1}^{N_{\mathrm{spb}}} 
     \frac{1}{N_{\mathrm{px}}}\sum_{p=1}^{N_{\mathrm{px}}}\lVert {\bf f}^{s,i}_p - {\bf f}^{\mathrm{gt}}_p \rVert_\epsilon \; .
\end{equation}
Here,  ${\bf f}^{s,i}_p$ denotes the flow estimate at scale $s$ and iteration $i$ upsampled to the original resolution and ${\bf f}^{\mathrm{gt}}_p$ denotes the ground truth flow -- both at pixel $p$. Furthermore,
$\|\cdot\|_\epsilon=\sqrt{\lVert\cdot\rVert_2^2+{\epsilon}}$
stands for the regularised $L_2$-norm 
with $\epsilon=1\mathrm{e}{-5}$, $N_{\mathrm{spb}}$ represents the number of samples per batch and $N_{\mathrm{px}}$ indicates the number of pixels in each sample.

Generalizing the original weighting scheme of of RAFT  
to multiple scales, we propose to compute the corresponding weights at scale $s$ and iteration $i$ as $\gamma_{s,i}=\gamma^{I_{\mathrm{tot}} - I_{s,i}}$ where the total number of iterations across all scales is given by
$I_{\mathrm{tot}}=\sum_{s=1}^{N_{\mathrm{scales}}} N_{\mathrm{iters}}(s)$ and the current iteration number w.r.t.\ all scales can be computed as $I_{s,i}=\sum_{k=1}^{s-1} N_{\mathrm{iters}}(k)\! + \!i$. By choosing $\gamma\!=\!0.8$ this ensures not only exponentially increasing weights for later iterations but also for finer scales. 

Similar to \cite{Sun2018_PWC}, we apply a different loss for fine-tuning.
However, we do not further increase the pixel-wise robustness as in \cite{Sun2018_PWC}, but propose a novel sample-wise robust loss via
 \eqref{msmiloss:pre} 
\begin{equation}
     \label{msmiloss:fine}
     L_{s,i}^{\mathrm{fine}} = \frac{1}{N_{\mathrm{spb}}}\sum_{{\mathrm{m}=1}}^{N_{\mathrm{spb}}} 
     \left( \frac{1}{N_{\mathrm{px}}}\sum_{p=1}^{{N_{\mathrm{px}}}} \lVert{{\bf f}^{s,i}_p - {\bf f}^{\mathrm{gt}}_p}\rVert_\epsilon + \epsilon' \right)^{\!\!q\;\;}
\end{equation}
where $q \! = \! 0.7$ and $\epsilon' \! = \!  0.01$. 
While penalizing per-pixel deviations from the ground truth linearly 
already ensures pixel-wise robustness by down-weighting outlier pixels \cite{Huber2004_RobStat}, our modification additionally allows to down-weight the impact of an entire sample in the batch, if the sample contains too many out\-liers. In that sense it is similar to \cite{Sun2018_PWC,Ilg2017_FlowNet2} where difficult samples were explicitly excluded from training.

\section{Experiments}
\label{sec:experiments}
\textbf{Training.}
We trained our 
method using two NVIDIA A100 GPUs. To this end, we followed a procedure similar to RAFT.
First, we pre-trained the model on FlyingChairs~\cite{Dosovitskiy2015_FlowNet} (Chairs) for 100K iterations 
with batch size of 10 and on FlyingThings3D~\cite{Mayer2016_FTH} (Things)
for 100K with a batch size of 6. Then we fine-tuned our model for 100K 
using a batch size of 6
in a mixed setting on MPI Sintel~\cite{Butler2012_Sintel}, KITTI 2015~\cite{Menze2015_KITTI}, HD1K~\cite{Kondermann2016_HD1K} and Things with the same proportions suggested by RAFT. 
We also used the same patch size, optimizer and learning rate schedule as RAFT with maximum step size of 0.0004 (Chairs), 0.0002 (Things) and 0.0002 (mixed).
However, note that unlike  
RAFT and
\cite{Long2022,Xu2021_Flow1D,Zhang2021_Separable,Jiang2021_GMA,Jiang2021_SCV} we do not fine-tune additionally on individual datasets afterwards.

\smallskip

\noindent
\textbf{Our Setting.} Our final model 
has about 13.5M parameters\linebreak 
and 
consists of 3 scales each using a correlation pyramid with 2 look\-up levels.
The overhead compared to the 5.3M parame- ters 
of RAFT is 
due to our multi-scale U-Net-style feature extractor. 
For image features, the number of channels 
from coarsest to finest scale reads
256, 128 and 96, respectively. For context features, 
this number is 
256 for all scales. 
Regarding training, we set the number of recurrent iterations 
from coarsest to finest scale
to 4, 6 and 8 iterations, respectively. For evaluation, we used 10, 15 and 20 iterations for Sintel and 4, 8 and 24 iterations for KITTI.
These numbers were found empirically  
by evaluating our pre-trained model on the respective training datasets.
Moreover, similar to RAFT, 
we considered a warm start approach. 
The resulting inference time for Sintel and KITTI is about $0.3$ seconds each.

\smallskip

\noindent
\textbf{Results.} 
To compare our method to the literature,
we evaluated it on the 
MPI Sintel and the KITTI 2015 benchmark. The
results are listed in \cref{eval:tab:literature},
where boldfaced and underlined entries indicate best and second-best results, respectively.
As one can see, our method shows excellent generalization capabilities across both benchmarks. On the one hand, its consistently outperforms RAFT with accuracy gains up to 14.6\%. On the other hand, with R1 on Sintel Clean, R2 on Sintel Final and R3 on KITTI, it is always among the top 3 methods. 
Moreover, it is particularly accurate
in non-occluded regions: R1 on Sintel Clean and Final, R2 on KITTI.
This observation is confirmed by the computed flow fields in  \cref{fig:benchmark_results_images} that show
comparably more structural 
details. 
Finally, we also evaluated results 
without considering our sample-wise robust loss during fine-tuning. 
Not only the results for KITTI show that this loss plays a fundamental role in our mixed training that allows 
a good generalization without individual fine-tuning, but also applying it clearly improves the results  for Sintel Final.

\begin{table*}[t!]
    \caption{Results on Sintel (test) and KITTI (test) using AEPE (Sintel) and Fl (KITTI). Mat = matched. Unmat = unmatched.}
    
	\vspace{2mm}
	
	\label{eval:tab:literature}
	\centering
	\begin{tabular}{l r c >{\columncolor[gray]{0.95}[6pt][4pt]}cccccc c >{\columncolor[gray]{0.95}[6pt][4pt]}cccccc c >{\columncolor[gray]{0.95}[6pt][4pt]}cccc c}
		\toprule
		 &&\;& \multicolumn{6}{c}{Sintel Clean (test)} 
		 &\!\!\!\!& \multicolumn{6}{c}{Sintel Final (test)} 
		 &\!\!\!\!& \multicolumn{3}{c}{KITTI (test)}
		  \\\cline{4-8}\cline{11-15}\cline{18-20}\\[-8mm]
		  \\\cline{4-8}\cline{11-15}\cline{18-20}
		 \rowcolor[gray]{1.0} Method &&\!\!\!\!& {\;All\;}
		 &\!\!\!\!\!\!\!\!\!\!\!& {\;Mat\;}
		 &\!\!\!\!\!\!\!\!& \!\!\!\!\!\!\!\!\!\!{\;Unmat\;}\!\!\!\!\!\!\!\!\!\!&
	  	 &\!\!\!\!& {\;All\;}
		 &\!\!\!\!\!\!\!\!\!\!\!& {\;Mat\;}
		 &\!\!\!\!\!\!\!\!& \!\!\!\!\!\!\!\!\!\!{\;Unmat\;}\!\!\!\!\!\!\!\!\!\!&
		 &\!\!\!\!& {\;Fl-all\;}
		 &\!\!\!\!\!\!\!\!\!\!\!& {\;Fl-noc\;}
		\\
		\midrule
		Flow1D~\cite{Xu2021_Flow1D} & ICCV'21              &\!\!\!\!&
		2.24                          &\!\!\!\!\!\!\!\!\!\!\!&
		0.97                          &\!\!\!\!\!\!\!\!\!\!\!&
		12.59                         &&
		\!\!\!\!&
		3.81                          &\!\!\!\!\!\!\!\!\!\!\!&
		1.95                          &\!\!\!\!\!\!\!\!\!\!\!&
		18.95                         &&
		\!\!\!\!&
		6.27                          &\!\!\!\!\!\!\!\!\!\!\!&
		-                             
		\\
		DICL-Flow+~\cite{Wang2020_DICL} & \!\!\!\!NeurIPS'20       &\!\!\!\!&
		1.86                          &\!\!\!\!\!\!\!\!\!\!\!&
		0.74                          &\!\!\!\!\!\!\!\!\!\!\!&
		11.01                         &&
		\!\!\!\!&
		3.32                          &\!\!\!\!\!\!\!\!\!\!\!&
		1.59                          &\!\!\!\!\!\!\!\!\!\!\!&
		15.76                         &&
		\!\!\!\!&
		6.31                          &\!\!\!\!\!\!\!\!\!\!\!&
		-                             
		\\
		SCV~\cite{Jiang2021_SCV} & \!\!\!\!CVPR'21       &\!\!\!\!&
		1.72                          &\!\!\!\!\!\!\!\!\!\!\!&
		0.57                         &\!\!\!\!\!\!\!\!\!\!\!&
		11.08                         &&
		\!\!\!\!&
		3.60                          &\!\!\!\!\!\!\!\!\!\!\!&
		1.70                          &\!\!\!\!\!\!\!\!\!\!\!&
		19.14                         &&
		\!\!\!\!&
		6.17                          &\!\!\!\!\!\!\!\!\!\!\!&
		-                             
		\\
		RAFT~\cite{Teed2020_RAFT} & ECCV'20 &\!\!\!\!&
		1.61                          &\!\!\!\!\!\!\!\!\!\!\!&
		0.62                          &\!\!\!\!\!\!\!\!\!\!\!&
		\phantom{0}9.65               &&
		\!\!\!\!&
		2.86                          &\!\!\!\!\!\!\!\!\!\!\!&
		1.41                          &\!\!\!\!\!\!\!\!\!\!\!&
		14.68                         &&
		\!\!\!\!&
		5.10                          &\!\!\!\!\!\!\!\!\!\!\!&
		3.07                          
		\\
		Separable Flow~\cite{Zhang2021_Separable}\!\!\!\! & ICCV'21      &\!\!\!\!&
		1.50                          &\!\!\!\!\!\!\!\!\!\!\!&
		0.57                          &\!\!\!\!\!\!\!\!\!\!\!&
	    \phantom{0}9.08               &&
	    \!\!\!\!&
	    \underline{2.67}              &\!\!\!\!\!\!\!\!\!\!\!&
		1.28                          &\!\!\!\!\!\!\!\!\!\!\!&
	    \underline{14.01}             &&
		\!\!\!\!&
		{\bf 4.64}                    &\!\!\!\!\!\!\!\!\!\!\!&
		{\bf 2.78}                     
		\\
		RFPM~\cite{Long2022} & WACV'22 &\!\!\!\!&
		1.41                          &\!\!\!\!\!\!\!\!\!\!\!&
		0.49                          &\!\!\!\!\!\!\!\!\!\!\!&
		\phantom{0}8.88               &&
		\!\!\!\!&
		2.90                          &\!\!\!\!\!\!\!\!\!\!\!&
		1.33                          &\!\!\!\!\!\!\!\!\!\!\!&
		15.69                         &&
		\!\!\!\!&
		\underline{4.79}              &\!\!\!\!\!\!\!\!\!\!\!&
		-                          
		\\
	    GMA~\cite{Jiang2021_GMA} & ICCV'21                 &\!\!\!\!&
		\underline{1.39}              &\!\!\!\!\!\!\!\!\!\!\!&
		0.58                          &\!\!\!\!\!\!\!\!\!\!\!&
		{\bf \phantom{0}7.96}         &&
		\!\!\!\!&
		{\bf 2.47}                    &\!\!\!\!\!\!\!\!\!\!\!&
		\underline{1.24}              &\!\!\!\!\!\!\!\!\!\!\!&
		{\bf 12.50}                   &&
		\!\!\!\!&
		4.93                         &\!\!\!\!\!\!\!\!\!\!\!&
		2.90                          
		\\
		MS-RAFT, w/o rob. (ours)\!\!\!\!\!\!\!\!\!\!\!\!\!\!\!\! &              &\!\!\!\!&
		\underline{1.39}              &\!\!\!\!\!\!\!\!\!\!\!&
		{\bf 0.46}                    &\!\!\!\!\!\!\!\!\!\!\!&
		\phantom{0}8.89               &&
		\!\!\!\!&
		           2.79               &\!\!\!\!\!\!\!\!\!\!\!&
		1.25                          &\!\!\!\!\!\!\!\!\!\!\!&
		15.36                         &&
		\!\!\!\!&
	               5.35               &\!\!\!\!\!\!\!\!\!\!\!&
		           2.98              
		\\
		MS-RAFT (ours)\!\!\!\! &              &\!\!\!\!&
		{\bf 1.37}                    &\!\!\!\!\!\!\!\!\!\!\!&
		\underline{0.48}              &\!\!\!\!\!\!\!\!\!\!\!&
		\phantom{0}\underline{8.67}   &&
		\!\!\!\!&
		\underline{2.67}              &\!\!\!\!\!\!\!\!\!\!\!&
		{\bf 1.20}                    &\!\!\!\!\!\!\!\!\!\!\!&
		14.70                         &&
		\!\!\!\!&
		4.88                          &\!\!\!\!\!\!\!\!\!\!\!&
		\underline{2.80}              
		\\
		\midrule
		\multicolumn{2}{l}{Improvement over RAFT} &\!\!\!\!&
		+14.6\%                       &\!\!\!\!\!\!\!\!\!\!\!&
		+23.1\%                       &\!\!\!\!\!\!\!\!\!\!\!&
		+10.1\%                       &&
		\!\!\!\!&
		+6.6\%                        &\!\!\!\!\!\!\!\!\!\!\!&
		+14.6\%                       &\!\!\!\!\!\!\!\!\!\!\!&
		-0.1\%                        &&
		\!\!\!\!&
		+4.3\%                        &\!\!\!\!\!\!\!\!\!\!\!&
		+8.8\%                        
		\\
		\bottomrule
	\end{tabular}
\end{table*}

\smallskip

\noindent
\textbf{Ablations:}
We discuss different aspects of our architecture showing 
warm start results on Sintel (train) with models pre-trained on Chairs and Things; see \cref{eval:tab:ablation:pretrain}. 
The first ablation shows that simply considering single-scale RAFT on a finer resolution of $\frac{1}{4}\times(h, w)$ does not improve results, although we increased the number of lookup levels in the correlation pyramid accordingly from 4 to 5 -- to have a comparable non-local lookup. 
The ablation also shows that increasing the number of recurrent iterations for single-scale RAFT from 12 to 18 -- which is the overall number of
iterations of our multi-scale method -- hardly changes the results 
for both the original and the finer resolution. 
Evidently, without a proper initialization and without the use of multi-scale features it seems difficult 
to obtain improvements.
For the single-scale variants we followed the same training and evaluation procedure
as RAFT. 

The remaining three ablations address aspects of 
our multi-scale method with finest scale  
$\frac{1}{4}\times(h, w)$. In our second ablation, we investigate the interplay between the number of coarse-to-fine scales and the number of lookup levels in the correlation pyramid. Thereby we chose the number of recurrent iterations during training such that the sum is always equal to 18 (2 scales: 8+10, 3 scales: 4+6+8, 4 scales: 3+3+5+7) and increased this number during evaluation according to RAFT.
The results show that our 3 scale approach using 2 lookup levels is the best compromise between Sintel Clean and Sintel Final. In our third ablation, we analyze the impact of our U-Net-style features compared to standard features 
(intermediate features in \cref{fig:unet} without enhancement). 
While our U-Net-style features show slightly worse results for Sintel Clean, they allow larger improvements for the more challenging final pass. 
Finally, we compare our multi-scale 
multi-iteration
loss to the same loss only applied to the finest scale.
From the results the clear 
advantage becomes explicit.

\begin{table}[t!]
	\caption{Ablation study. Warm start results on Sintel (train).}
	\vspace{2mm}
	
	\label{eval:tab:ablation:pretrain}
	\centering
	\begin{tabular}{lccccc}
		\toprule
		Pre-trained on Chairs \& Things & & \multicolumn{3}{c}{Sintel (train)}
		\\\cline{3-5}
		& & {\;Clean\;} && {\;Final\;} &
		\\
		 \midrule
		\multicolumn{6}{l}{{\textbf{\!\!\! 1. Single-Scale RAFT:} Finest Scale / Recurrent Iterations}}\\
		\midrule
		$\frac{1}{8}\times$(h,w), 12 iter (RAFT warm) \!\!\!\!\!\!\! 
		 &&
		{\bf 1.40} && 
		\bf{2.67} &
		\\[0.5mm]
		$\frac{1}{8}\times$(h,w), 18 iter\; &&
		\underline{1.45} &&
		\underline{2.70} &
		\\
		\midrule
		$\frac{1}{4}\times$(h,w), 12 iter\; &&
		1.58 &&
		3.10 &
		\\[0.5mm]
		$\frac{1}{4}\times$(h,w), 18 iter\; &&
		1.52 &&
		3.08 &
		\\

        \midrule
        
        \multicolumn{6}{l}{{\textbf{\!\!\! 2. Multi-Scale RAFT:} Resolution Scales / Look-Up Levels\!\!\!}}
		\\
		\midrule
        
		2 scales / 3 levels \; &&
		1.16 &&
		3.07 &
		\\
		2 scales / 4 levels \; &&
		1.14 &&
		2.64 &
		\\
		2 scales / 5 levels \; &&
		\bf 1.11 &&
		2.66 &
		\\ 
		\midrule
		{3 scales / 2 levels (ours)} \; &&
		{\underline{1.13}} &&
		\underline{2.60} &
		\\
		3 scales / 3 levels \; &&
		1.15 &&
		2.66 &
		\\
		3 scales / 4 levels \; &&
		1.14 &&
		2.70 &
		\\
			\midrule
		4 scales / 1 levels \; &&
		1.19 &&
		{\bf 2.53} &
		\\
		4 scales / 2 levels \; &&
		1.16 &&
		2.67 &
		\\
		4 scales / 3 levels \; &&
		1.22 &&
		2.67 &
		\\
		
		\toprule
        \multicolumn{6}{l}{\textbf{\!\!\! 3. Multi-Scale Features:} U-Net-style vs.\ Standard}
		\\
		\midrule
		{U-Net-style (ours)} \; &&
			1.13 &&
		{\bf 2.60} &
		\\
        Standard \; &&
        {\bf 1.11}\ &&
		2.68 &
		\\
		
		\toprule
        \multicolumn{6}{l}{\textbf{\!\!\! 4. Multi-Scale Loss:} Single-Scale vs.\ Multi-Scale}
		\\
		\midrule
		{Multi-scale loss (ours)} \; &&
		{\bf 1.13} &&
		{\bf 2.60} &
		\\
        Single-scale loss \; &&
        2.26 &&
		4.09 &
		\\
		\bottomrule
	\end{tabular}
\end{table}

\section{Conclusion}
\label{sec:conclusion}
By combining different hierarchical concepts within a single architecture, we showed that multi-scale ideas are still valuable in optical flow estimation. While coarse-to-fine schemes enabled the use of finer scales by providing helpful initializations, correlation pyramids introduced non-local cost information at each scale.
Moreover, semantically enhanced multi-scale features allowed a more accurate and more robust matching. Finally, our sample-wise robust multi-scale multi-iteration loss provided good generalization capabilities\linebreak without requiring individual fine-tuning.
Experiments showed state-of-the-art results -- especially in non-occluded regions.

\bibliographystyle{IEEEbib}
\bibliography{refs}

\end{document}